# A New Hybrid metric for verifying parallel Corpora of Arabic-English

Saad Alkahtani, Wei Liu, and William J. Teahan

School of Computer Science, Bangor University, Bangor, United Kingdom
{s.alkahtani,w.liu,w.j.teahan}@bangor.ac.uk

## ABSTRACT

*This paper discusses a new metric that has been applied to verify the quality in translation between sentence pairs in parallel corpora of Arabic-English. This metric combines two techniques, one based on sentence length and the other based on compression code length. Experiments on sample test parallel Arabic-English corpora indicate the combination of these two techniques improves accuracy of the identification of satisfactory and unsatisfactory sentence pairs compared to sentence length and compression code length alone. The new method proposed in this research is effective at filtering noise and reducing mis-translations resulting in greatly improved quality.*

## 1. BACKGROUND AND MOTIVATION

The history of translation between natural languages can be traced back to the beginning of human culture, with its major mission being to expand the informativeness of one language, decrease the misunderstanding in dialogue, and even contribute to the growth of cultures [5]. Language translation is seen as a valuable social science oriented industry to help people develop international relationships.

Early pioneering machine translation systems were developed in the 1950s and 1960s [8]. Machine translation requires the development of computing technologies in the areas of computational linguistics and natural language processing. Machine and machine-aided translation are gaining in accuracy and popularity. Computer technology is essential to manage the large amounts of text available that may need to be translated.

Our research specifically explores the development of resources for Arabic-English translation, two important global languages. Arabic is spoken throughout the world and is the official language of 27 states, the third most after French and English. The term 'Arabic' when used in this paper refers to a variety of dialects belonging to the Central Semitic languages. English is the essential language of commerce and science and is the common lingua franca between many nations. Both languages are two of the six official languages of the United Nations.

Arabic is the primary language for over 380 million native Arabic speakers worldwide [15]. Computing research and applications designed for Arabic has increased in recent years. In particular, there have been a significant number of people accessing the Internet in Arabic. The majority of these users do not speak any language other than Arabic, which means they cannot easily access the vast variety of English information available. At the same time, global interest in Arabic countries, in culture, politics, economics, and other areas has expanded worldwide.

Language corpora have become increasingly important in natural language processing, and machine translation in particular. Corpora are an important resource often used for training purposes for statistical-based language modelling and machine translation. Large-scale parallel corpora are needed to construct statistical machine translation systems. Given the large number of Arabic speakers and the global importance of English, it is vital that translation between these languages be facilitated by the use of high quality parallel corpora. However, the structural differences between these languages present a challenge for machine translation. Arabic requires an altogether different treatment than European languages because of its unique morphology. Arabic and English are also different in a number of graphology aspects as Table 1 shows.

Table 1. A list of differences between the Arabic and English languages.

| Graphology Aspects | Arabic Language | English Language |
| --- | --- | --- |
| **Written and Read** | From right to left | From left to right |
| **Capitalization** | No | Yes |
| **Size of Alphabet** | 28 letters | 26 letters |
| **Gender Differentiation** | Verbs and sentence structures | No differentiation |
| **Types of Sentences** | Nominal and verbal | Verbal |
| **Plural Forms** | Singular, dual and plural | Singular and plural |
| **Position of Adjective** | After the noun | Before the noun |

The use of parallel Arabic-English corpora to train statistical MT models provides an effective way for building MT systems. However, Arabic-English parallel texts of high quality are still very limited and are not available in satisfactory quantities, therefore most translations are performed manually, a time consuming and often error-filled process. Limitations of existing parallel corpora include incomplete data, untagged entries, with only limited text genres being available (such as news stories). In addition, many of the better quality corpora are not available for public use with fees in the thousands of dollars. For example, a list of corpora that were available from the Linguistic Data Corporation (LDC) in 2013 at the beginning of our research project is shown in Table 2 [12]. These costs are often unaffordable for most students, and also for many researchers or small research groups.

Table 2. Parallel Arabic-English Corpora as provided by the LDC in 2013 [12].

| Corpora | Size (Words) | Price (US $) |
| --- | --- | --- |
| ISI Arabic-English Automatically Extracted Parallel Text | 31M | $4000 |
| Uma Arabic English Parallel News Text | 2M | $3000 |
| Arabic-English Parallel Translation | 42K | $3000 |
| Multiple Translation Arabic (Part 1) | 23K | $1000 |
| Multiple Translation Arabic (Part 2) | 15K | $1000 |
| Arabic Newswire English Translation Collection | 551K | $1500 |
| Arabic News Translation Text (Part 1) | 441K | $3000 |
| GALE Phase 1 Arabic Broadcast News Parallel Text (Part 1) | 90K | $1500 |
| IGALE Phase 1 Arabic Broadcast News Parallel Text (Part 2) | 56K | $1500 |
| UN Bidirectional Multilingual | 1M | $4000 |

Another motivation behind our research is to develop techniques that would allow the construction of high quality parallel corpora that are free for everyone to use by improving the quality of the data as well as combining existing corpora, and by constructing much larger corpora, for example by using web scraping techniques. In order to achieve this task, we believe that a more accurate and robust metric than existing methods (such as sentence length) is needed for matching sentence pairs between languages.

This paper is organised as follows. In the next section, we review some of the work that is related to the present work. Note that not all of the related work has been included (especially for the use of sentence length as a metric for alignment) due to the many publications in this area. In section 3, the new hybrid metric is described. The experimental evaluation is described in section 4, with the conclusion in the final section.

## 2. RELATED WORK

In a parallel bilingual corpus, textual elements (e.g. paragraphs, sentences, phrases, words) alignment is an essential job for statistical machine translation. There have been a number of different approaches for sentence alignment such as sentence length, word co-occurrence, cognates, dictionaries and parts of speech etc. for a parallel bilingual corpus [13].

The sentence length metric assumes that the length for each sentence will be kept the same when it is translated from the source language into the target language. Gale and Church [6] aligned parallel sentences in English-French and English-German corpora based on a sentence length metric that required calculating the character length of all sentences. Gale and Church's [6] overall accuracies were 97% for English-German and 94% for English-French. Wu [21] aligned English-Chinese corpora by using sentence length values and reached an accuracy of 95%. Kay and Röscheisen [9] developed a program that combined word and sentence alignment and calculated the word probabilities by using the dice co-efficient. Haruno and Yamazaki [7] used a similar method plus a bilingual dictionary for aligning English-Japanese corpora. Papageorgiou, Cranias, and Piperidis [17] used the sentence alignment metric based on the highest matching part of speech tags and matches restricted to nouns, adjectives and verbs, and reached 99% accuracy. Simard, Foster and Isabelle [19] used cognate-based approaches and found that sentence length difference worked well for sentence alignment. However, Melamed [14] pointed out that because results were only reported for a relatively easy bilingual text, comparing two algorithms' performances in the literature is difficult. In addition, Brown *et al*. [3] calculated sentence length by using the number of words instead of the number of bytes or characters, which generated similar accuracies between 96% and 97%.

In the last decade, there have been few new proposals for sentence alignment for parallel bilingual corpora [22]. One disadvantage of existing sentence alignment algorithms is that it is less effective when linking corresponding sentences if they are one-to-many or many-to-one mutual translations [11].

Our approach, as well as using sentence length, also makes use of the strong correlation in compression code length (number of bits required to encode the text) between original sentences and accurately translated sentences. We show in this paper that this correlation can be used to evaluate and improve the quality of bilingual parallel corpora. If encoded into bit strings, almost all natural language text contain redundant bits that can be removed without affecting the information they carry. An observation by Behr *et al*. on natural languages indicates that all natural languages have similar cross entropies [2]. According to Shannon's information theory [18], a derived hypothesis from this observation will be that all natural languages can be encoded into the same length of bit strings for the same information if redundant bits are excluded.

Our work with using compression code length metrics for sentence alignment with Chinese-English corpora have shown they can be very effective [13]. Our idea of using compression code length as a sentence alignment metric hinges on the premise that the compression of co-translated text (i.e. documents, paragraphs, sentences, clauses, phrases) should have similar code lengths [2]. This is based on the notion that the information contained in the co-translations will be similar. Since compression can be used to measure the information content, we can simply look at the ratio of the compression code lengths of the co-translated text pair to determine whether the text is aligned. That is, if you have a text string (i.e. document, paragraph, sentence, clause or phrase) in one language, and its translation in another language, then the ratio of the compression code lengths of the text string pair should be close to 1.0. This approach and the new hybrid approach are described in more detail in the next section.

## 3. A New Metric for Checking the Quality of Parallel Sentence Pairs

In this section, we describe how we use a distance metric based both on sentence lengths and on compression code lengths (using the Prediction by Partial Match (PPM) compression scheme) in order to check the quality of the sentence pairs.

### 3.1. PPM Compression Code Length Metric

The Prediction by Partial Matching (PPM) compression scheme, first proposed by Cleary and Witten in 1984 [4], predicts the next symbol or character from a fixed order context. The context models are adaptively updated as the text is processed sequentially using an online adaptive process. For both Arabic and English text [1, 9], experiments have shown that order 5 models (using fixed order contexts of length 5) perform best at compressing the text using the PPMD variant of the algorithm developed by Howard in 1993 based on the PPMC variant devised by Moffat [20]. The main difference between PPMC and PPMD (and other variants PPMA and PPMB) is the calculation of the *escape* probability when the model needs to back off to lower order models if a symbol is not predicted by a higher order model.

Formally, the estimation of the *escape* probability for PPMD is $e = t_d/2n$ and for the symbol probability is $p(\varphi) = (2c(\varphi) - 1)/2T_d$, where: $d$ is the current coding order of a model; $\varphi$ is an upcoming symbol ($\varphi = x_{n+1} \in A$); $s_d$ is the current context $s_d = x_n, \cdots, x_{n-d+1}$; $c_d(\varphi)$ is the number of times that the symbol $\varphi$ in the context $s_d$; $t_d$ is the total number of unique symbols that occur after the context $s_d$; $T_d$ is the total number of times that the context $s_d$ has been seen with $T_d = \sum c_d(\varphi)$; $e$ is the *escape* probability; and $p(\varphi)$ is the probably of the upcoming symbol $\varphi$.

In this paper for our experiments, we use PPMD with $d = 5$ since as stated, experience shows that this is most effective for compressing English and Arabic text and performs better than PPMA, PPMB and PPMC. Table 3 shows how the probabilities are estimated using PPMD when the model has been trained on the sample text string "سبييللسلسلسبيلا". The table shows the predictions, frequency counts $c$ and probability estimates $p$ for the order $k = 3$, $k = 2$, $k = 1$, $k = 0$ and $k = -1$ PPMD contexts (where $k$ is the order of the model or context length). For example, only one symbol has been predicted in the single order 3 context – this has occurred once in the training text, and therefore its probability estimate that it will occur again is 3/4 and the probability estimate that a previously unseen symbol in this context will occur instead is 1/4 therefore necessitating the use of lower order models in order to estimate the probability of the unseen symbol. The model will keep on escaping down until it encounters a context where the symbol has been seen before or the symbol will be encoded using the default $k = -1$ context where every symbol is estimated with equal probability $1/|A|$ where $|A|$ is the size of the alphabet.

Table 3. PPMD order 3 model after processing the text string "سبيلللسلسلسبيلا".

| Order k = 3 | | | Order k = 2 | | | Order k = 1 | | | Order k = 0 | | |
|---|---|---|---|---|---|---|---|---|---|---|---|
| Prediction | c | p | Prediction | c | p | Prediction | c | p | Prediction | c | p |
| ل→سبي | 2 | 3/4 | ي→سب | 2 | 3/4 | ب→س | 2 | 3/8 | →س | 4 | 7/30 |
|  |  |  |  |  |  |  |  |  | →ب | 2 | 3/30 |
|  |  |  |  |  |  |  |  |  | →ي | 2 | 3/30 |
|  |  |  |  |  |  | →ل | 2 | 3/8 | →ل | 6 | 11/30 |
|  |  |  |  |  |  |  |  |  | →ا | 1 | 1/30 |
| → esc | 1 | 1/4 | → esc | 1 | 1/4 | → esc | 2 | 2/8 | → esc | 5 | 5/30 |
| ل→يل | 1 | 1/4 | ل→يي | 2 | 3/4 | ي→ب | 2 | 3/4 | Order k = −1 | | |
| →ا | 1 | 1/4 |  |  |  |  |  |  | Prediction | c | p |
| → esc | 2 | 2/4 | → esc | 1 | 1/4 | → esc | 1 | 1/4 | → A | 1 | 1/\|A\| |
| ل→يلل | 1 | 1/2 | ل→يل | 1 | 1/4 | ل→ي | 2 | 3/4 |  |  |  |
|  |  |  | →ا | 1 | 1/4 |  |  |  |  |  |  |
| → esc | 1 | 1/2 | → esc | 2 | 2/4 | → esc | 1 | 1/4 |  |  |  |
| س→للل | 1 | 1/2 | ل→ال | 1 | 1/4 | ل→ل | 2 | 3/12 |  |  |  |
|  |  |  | →ا | 1 | 1/4 | →س | 3 | 5/12 |  |  |  |
|  |  |  |  |  |  | →ا | 1 | 1/12 |  |  |  |
| → esc | 1 | 1/2 | → esc | 2 | 2/4 | → esc | 3 | 3/12 |  |  |  |
| ل→للس | 1 | 1/2 | ل→لس | 2 | 3/6 |  |  |  |  |  |  |
|  |  |  | →ب | 1 | 1/6 |  |  |  |  |  |  |
| → esc | 1 | 1/2 | → esc | 2 | 2/6 |  |  |  |  |  |  |
| س→لسل | 2 | 3/4 | س→سل | 2 | 3/4 |  |  |  |  |  |  |
| → esc | 1 | 1/4 | → esc | 1 | 1/4 |  |  |  |  |  |  |
| ل→سلس | 1 | 1/4 |  |  |  |  |  |  |  |  |  |
| →ب | 1 | 1/4 |  |  |  |  |  |  |  |  |  |
| → esc | 2 | 2/4 |  |  |  |  |  |  |  |  |  |
| ي→لسب | 1 | 1/2 |  |  |  |  |  |  |  |  |  |
| → esc | 1 | 1/2 |  |  |  |  |  |  |  |  |  |

### 3.2. Code Length Ratio Distance Metric for Matching Sentences

The term code length refers to the size (in bytes) of the compressed output file produced by the PPM compression algorithm. When using PPM to compress Arabic or English text, the code length is a measure of the cross-entropy of the text, which is the average size (in bytes) per character for the compressed output string. Theoretically, the cross-entropy is estimated as follows:

$$H(S) = \frac{1}{k}\log_2 p(S) = -\frac{1}{k}\sum_{i-1}^{k} -\log_2 p(x_i|x_1,\cdots,x_{k-1})$$

where $H(S)$ is the average number of bits to encode the text and $k$ is the order of the model (e.g. 5 for the models used in this paper).

Note that the compression code length, the number of bits required to encode the text string losslessly, so that it can be unambiguously decoded, can be expressed simply as $nH(S_L)$.

The ratio of the compression code lengths of the parallel text strings for languages $E$ (English) and $A$ (Arabic) is defined as follows:

$$R(S^E, S^A) = \frac{n}{m} \times \frac{H(S^E)}{H(S^A)}$$

where $S^E$ is an English text string with length $n$ and $S^A$ is an Arabic text string with length $m$.

The code length ratio (*CR*) is defined as:

$$CR = max\{R^{E,A}, R^{A,E}\}$$

Liu *et al.* [13] have shown that *CR* is a more effective distance metric for sentence alignment of Chinese-English parallel corpora than a distance metric based on sentence length. A primary purpose of the research reported in this paper was to investigate whether this would also be the case for Arabic-English parallel corpora.

### 3.3. Sentence Length Ratio Distance Metrics for Matching Sentences

Automatically generated bilingual corpora often have a large number of noisy sentence pairs. Consequently, researchers have devised various methods to filter noisy sentences from parallel corpora [10]. However, for our experiments discussed in Section 4, we have found a new technique based on a combination of the compression Code Length Ratio (*CR*) described above and the standard Sentence Length Ratio (*SLR*) described by Mújdricza-Maydt [16] is the most effective for Arabic-English sentence pairs in order to achieve a high-quality corpus as a result.

The Sentence Length Ratio (*SLR*) for a pair of translation sentences for Arabic and English can be calculated as follows:

$$SLR = max\left\{\frac{L^A}{L^E}, \frac{L^E}{L^A}\right\}$$

where $L^A$ is the length of the text for Language *A*.

## 4. EXPERIMENTAL EVALUATION

### 4.1. Developing the Test Corpora

For our experimental evaluation, two parallel Arabic-English test corpora were created. A large corpus (Corpus A) was first created containing fifty-eight million words that was collected from two online sources Al Hayat (http://www.alhayat.com) and OPUS (http://opus.lingfil.uu.se) with permission obtained from the owners of the data. OPUS is an open source parallel corpus that provides a large collection of translated texts from the web. All the online data was collected automatically and as a result the original texts are not of high quality. However, a primary purpose of our research is to develop a more reliable collection based on this and other data with poor quality translations filtered out using our sentence matching metrics.

A second much smaller test corpus (Corpus B) was created containing 10,000 translations judged satisfactory and 2,000 translations judged unsatisfactory. These were manually selected from Corpus A and formed the ground truth data for our experiment.

The files in Corpus A were also classified into 13 categories such as Books, Business, Cinema, Conferences, Crimes, Decisions, Economy, Geographies, Issues, Law, Politics, Reports and Stories as described in Table 4. The number of Arabic and English characters and words in each of the categories are also shown in the table. In total, 58,380,784 words were collected comprising 27,775,663 Arabic words and 30,808,480 English words.

Table 4. Character and word counts for test Corpus A.

| Categories | Arabic Characters | English Characters | Arabic Words | English Words |
|---|---|---|---|---|
| Books | 10,574,252 | 7,242,426 | 931,836 | 1,079,699 |
| Business | 26,367,126 | 17,987,925 | 2,289,276 | 2,624,274 |
| Cinema | 61,557,926 | 36,482,892 | 7,919,902 | 8,127,509 |
| Conferences | 21,696,083 | 15,129,972 | 1,879,527 | 2,215,857 |
| Crimes | 10,147,866 | 6,473,170 | 933,842 | 1,005,221 |
| Decisions | 15,863,975 | 10,822,315 | 1,397,181 | 1,605,851 |
| Economy | 25,962,438 | 17,760,514 | 2,266,424 | 2,599,651 |
| Geographies | 16,096,053 | 10,924,063 | 1,392,099 | 1,595,115 |
| Issues | 11,390,107 | 6,937,792 | 1,051,195 | 1,042,316 |
| Law | 16,083,105 | 10,936,231 | 1,407,292 | 1,597,873 |
| Politics | 23,427,958 | 15,675,917 | 2,035,969 | 2,304,233 |
| Reports | 15,960,285 | 10,819,195 | 1,388,457 | 1,590,056 |
| Stories | 29,703,105 | 20,294,105 | 2,882,663 | 3,420,825 |
| **Total** | 284,830,279 | 187,486,517 | 27,775,663 | 30,808,480 |

## 4.2. Compression Experiments

Preliminary compression experiments were conducted to determine if the *CR* compression code length measure would be effective as a metric for measuring the quality of translation between sentence pairs of Arabic and English.

Standard PPM is an adaptive technique with its language models starting from null when the beginning of a text string is processed. The context frequency counts from which the probability estimates are made are then updated as the text string is processed sequentially. For longer text strings (such as documents and paragraphs), the PPM algorithm will usually have enough text in order to train its models effectively so that higher order contexts are being used for most predictions with less need to escape down to lower order contexts.

One obvious concern when using PPM code lengths is that sentences may not be long enough in order that more reliable probability estimates can be made for the *CR* calculation. A simple expedient in overcoming this difficulty is to prime the PPM models prior to the compression. We can use a large corpus that is representative of the language (such as the Brown corpus for English and the CCA corpus for Arabic) in order to prime the models prior to the compression being performed (i.e. 'train' the models using the priming text). This approach has been found to be very effective, for example, when using compression code length based metrics for sentence alignment between Chinese and English [13].

The purpose of the preliminary experiments described in this section were to determine how effective priming of the PPM models was for compressing Arabic sentences, and also how effective the primed PPM compression method as a sentence matching metric. A key requirement of using the *CR* metric is that the compression code lengths in the two different languages should be the same for sentences that are co-translations of each other. The intuition is that if the sentences are satisfactory co-translations, then they should convey exactly the same amount of information. Since compression code length is an effective method for measuring information (see [20] for several references), then we would expect that roughly 50% of the compression code lengths of sentences in one language to be longer than compression code lengths of sentences in the other language, and vice versa.

Clearly, this correlation would not be expected for sentence lengths. It is quite common that English sentences are shorter than their co-translation counterparts other languages (although this is not the case when compared with the Arabic sentences as reported below in this section). However, this should not be the case for compression code lengths if our intuition about the correlation between information is correct. If we find that the compression code lengths do not correlate, then the reason for this is more likely to be as a result of a less effective compression algorithm being used for one language resulting in a less accurate estimate of the information contained in the sentence.

In a preliminary experiment, 10 sample sentence pairs in Arabic and English were randomly chosen from Corpus A. The 10 sample sentence pairs that we used are shown in Table 5.

Table 5. Sample sentence pairs that were used in the initial compression experiments.

| Sent. ID | Arabic Sentence | English Sentence |
|---|---|---|
| 1 | موضوعي اليوم جديٌّ ولكن أبدأه بطرفة قديمة استدراجا للقارئ. | My topic today is a serious one, but I will begin with an old anecdote, to lure the reader in. |
| 2 | الوقوف في الجانب الصحيح من التاريخ محاولة لتبرير الحروب العادلة. | Standing on the right side of history represents an attempt to justify just wars. |
| 3 | كنت أهازرها إلا أنها فكرت، وسألتني هل أعتقد حقاً أن البكاء وسيلة أفضل لكسب الأصوات. | I was joking with her, but she took it seriously and asked me whether I really believed that crying was a better way to win votes. |
| 4 | هكذا الدنيا، جُنازِة أو جُوازٍ كما يقول اللبنانيون. | Such is life, a wedding or a funeral, as the Lebanese say. |
| 5 | هذا الرجل يقول: إنه يعرف ما لا يعرف قضاة لجنة الانتخابات | This man is saying that he knows something the judges on the Election Commission do not know. |
| 6 | فلندع مجدداً ريادتنا الربيع، ونحصي الخيبات، ومرارات صيفٍ يائس. | So let us once again claim to be the precursors of the spring, count the disappointments and tally the bitterness of a wretched summer. |
| 7 | وأن الذين توجهوا بعدها إلى القصر تصورا أن الرجل يجلس خلفه في انتظارهم! | Those who subsequently headed to the palace, truly imagined that the man was sitting there, waiting for them! |
| 8 | هو أخيراً ارتاح، بعد رحلة الآلام والآمال والنكبات والانتصارات، وترك لنا جميعاً مثلاً يُحتذى. | He has finally rested, after a journey of pains, hopes, disasters and triumphs, and left us all an example to be followed. |
| 9 | سيكون هناك شيء جديد تسمعه. | You will have something new to listen to it. |
| 10 | العسكريون أكثر تمسكاً بالدولة المدنية الديموقراطية والعلمانية. | The militaries are not more persistent on the civil, democratic and secular state. |

The results of compressing these sentences using the PPM compression scheme are shown in Table 6. The table lists the number of bytes that various variants of PPM produced as compressed output. For example, for sentence pair with id 1 (i.e. the first in Table 5), the WOT variant required 69 bytes to compress the Arabic sentence, compared to 69 bytes to compress the English sentence. In contrast, the sentence lengths are very different – the Arabic sentence is 59 characters (bytes) long compared to 95 characters for the English sentence.

Order 5 PPMD (as described above in Section 3.1) was used for these experiments. The WOT variant is for PPM without priming (i.e. no training). The WT variant is for PPM with priming. In this case, the PPM model was trained on the Brown Corpus prior to compressing the English text, whereas the PPM model was trained on the CCA Corpus prior to compressing the Arabic text. The WTPP variant used the same priming approach as for the WT variant, but also adopted a pre-processing algorithm to convert the UTF-8 encoded Arabic text into a number string before it was compressed by the PPMD5 compressor. This approach is described in detail in [1]. This leads to significantly better compression as a result for Arabic text and therefore leads to a better estimate of the information contained in the Arabic sentence.

Table 6. Compression results of the sample sentences. The PPMD5 compression code length results list the size in bytes of the compressed output produced by various variants of the PPMD5 compressor.

| Sentence ID | Sentence Length | | PPMD5 Compression Code Length | | | | | |
| --- | --- | --- | --- | --- | --- | --- | --- | --- |
| | | | (WOT) | | (WT) | | (WTPP) | |
| | Arabic | English | Arabic | English | Arabic | English | Arabic | English |
| 1 | 59 | 95 | 69 | 69 | 32 | 29 | 26 | 29 |
| 2 | 68 | 82 | 62 | 62 | 31 | 22 | 24 | 22 |
| 3 | 84 | 132 | 83 | 88 | 41 | 35 | 31 | 35 |
| 4 | 53 | 59 | 55 | 50 | 32 | 19 | 27 | 19 |
| 5 | 58 | 94 | 59 | 64 | 25 | 24 | 20 | 24 |
| 6 | 67 | 136 | 70 | 93 | 39 | 37 | 31 | 37 |
| 7 | 72 | 110 | 72 | 76 | 33 | 30 | 26 | 30 |
| 8 | 93 | 123 | 87 | 86 | 43 | 37 | 35 | 37 |
| 9 | 27 | 45 | 36 | 38 | 13 | 14 | 11 | 14 |
| 10 | 63 | 83 | 61 | 61 | 28 | 23 | 20 | 23 |

From the table, we can see there is a clear mis-match as expected between the Arabic and English sentence lengths. This provides clear evidence that metrics based on techniques well founded in information theory (as is the case for compression code length based metrics) have merit since they lead to better correlation.

The WOT variant does a surprisingly good job of matching the sentences with the Arabic bytes size being close to the English bytes size. However, again in most cases, the number of bytes of the compressed English output is greater than the number of bytes of the compressed Arabic output. For the WT variant, the opposite story is now the case – the compressed Arabic bytes is now usually greater the compressed English bytes. This indicates that the compression method being used for the Arabic text is probably not as well tuned as is the case for the English scheme (since the use of PPM for compressing English text has been fine-tuned over many years [20]). This problem was addressed in recent research on the compression of Arabic text [1] where it was found that using pre-processing techniques significantly improves PPM-based compression for Arabic in many cases by over 25%. When we apply these techniques (i.e. this is what we call the WTPP variant), then a more desired set of mixed results is achieved (where code lengths are sometimes greater for Arabic and sometimes greater for English).

In order to investigate this further and to confirm whether we have a compression method for Arabic text that produces compression code lengths that correlate well with those produced by the compression method being used for English text, we conducted further experiments using the WTPP PPM variant on the whole of the test Corpus A and in each of the categories as well. The results are listed in Table 7. The percentage of sentence pairs for which the Arabic sentence lengths are greater than for their English counterparts is shown in column 2. For example, for the Books category, it was found that only 8.55 % of the Arabic sentences are longer. In contrast, the results in the third column, which lists the percentage of sentence pairs for which the Arabic compression code lengths are greater than for their English counterparts, show that the comparison is more even, with most results being around the 50% mark, except for the

Crimes category with 62.48%. Due to this result, the nature of the sentences in this category should be investigated further.

Table 7. Percentage of Arabic sentences lengths or compression code lengths greater than their English sentence counterparts for the test Corpus A.

| Categories | % of Arabic sentence lengths that are greater | % of Arabic compression code lengths that are greater |
|---|---|---|
| **Books** | 8.55 | 54.96 |
| **Business** | 16.58 | 56.93 |
| **Cinema** | 35.43 | 44.94 |
| **Conferences** | 17.09 | 56.26 |
| **Crimes** | 21.88 | 62.48 |
| **Decisions** | 7.93 | 52.74 |
| **Economy** | 16.80 | 57.02 |
| **Geographies** | 14.79 | 58.97 |
| **Issues** | 16.30 | 53.71 |
| **Law** | 15.40 | 56.73 |
| **Politics** | 16.23 | 55.25 |
| **Reports** | 16.53 | 58.06 |
| **Stories** | 11.77 | 48.79 |
| **Average** | 16.56 | 55.14 |

These results provide reassuring evidence that the compression methods we have adopted produce the desired (and necessary) correlated data for the subsequent experiments we conducted that are described in the next section.

### 4.3. Analysing the quality of translations in the test corpora

Experiments were performed using the ground truth data in Corpus B to determine the best thresholds and combinations for the *CR* and *SLR* metrics in order to accurately filter out the unsatisfactory translations. For the *CR* calculations listed there, the WTPP PPMD5 variant (as stated, which was primed on the CCA corpus) was used to compress the Arabic text sentences, whereas standard PPMD5 primed on the Brown corpus was used for the English text sentences.

Various thresholds were applied firstly using *SLR* by itself, secondly using *CR* by itself, and thirdly by applying the same threshold to both *SLR* and *CR* together. If the calculated distance metric exceeded the threshold value(s), then the translation sentence pair was judged to be unsatisfactory, otherwise it was judged to be satisfactory.

The results of how accurate the filtering process was against the ground truth data in test Corpus B are shown in Table 8. The table shows the threshold values that were used for both the *SLR* and *CR* calculations in the leftmost column. The accuracy results are then provided in the subsequent columns. (This is the percentage of correct classifications made by the *SLR*, *CR* or *SLR&CR* metrics where a correct classification is made when the metric at a specific threshold judges the sentence pair to be satisfactory or unsatisfactory and this matches the ground truth judgment). The results are split for the satisfactory and unsatisfactory sentence pairs, with the average results provided in the final columns.

Table 8 shows, for example, that *SLR* with a threshold of 2.5 or higher is able to accurately classify 100% of the satisfactory translations whereas the threshold where this occurs for *CR* is at 2.25. For the unsatisfactory translations, 100% of these will be identified using *SLR* if the threshold is set at 1.5 or less, whereas the highest accuracy for *CR* is 97.45% when the threshold is set as low as 1.25 (meaning most sentence pairs will be rejected). The only calculation that results in an average accuracy of 100% for all sentence pairs (both satisfactory and unsatisfactory) occurs when both *SLR* and *CR* are combined together with a threshold of 2.5. Figure 1 shows the tendencies of the classification of the satisfactory translations and unsatisfactory translations for test Corpus B using different threshold values.

Table 8. Comparison of accuracies among different threshold values when using the different sentence matching metrics on test Corpus B.

| Threshold Values | 10000 Satisfactory Translations Accuracies (%) | | | 2000 Unsatisfactory Translations Accuracies (%) | | | Average Accuracies (%) | | |
|---|---|---|---|---|---|---|---|---|---|
| | *SLR* | *CR* | *SLR & CR* | *SLR* | *CR* | *SLR & CR* | *SLR* | *CR* | *SLR & CR* |
| 1.25 | 20.29 | 89.91 | 17.11 | 100 | 97.45 | 100 | 60.15 | 93.68 | 58.56 |
| 1.50 | 62.35 | 97.86 | 61.10 | 100 | 78.05 | 100 | 81.18 | 87.96 | 80.55 |
| 1.75 | 88.15 | 99.24 | 87.58 | 99.95 | 43.40 | 100 | 94.05 | 71.32 | 93.79 |
| 2.00 | 96.50 | 99.76 | 96.30 | 99.35 | 24.85 | 100 | 97.93 | 62.31 | 98.15 |
| 2.25 | 98.90 | 100 | 98.90 | 98.45 | 15.00 | 100 | 98.68 | 57.50 | 99.45 |
| 2.50 | 100 | 100 | 100 | 97.20 | 11.40 | 100 | 98.60 | 55.70 | 100 |
| 2.75 | 100 | 100 | 100 | 70.25 | 7.35 | 72.30 | 85.13 | 53.68 | 86.15 |
| 3.00 | 100 | 100 | 100 | 50.40 | 4.95 | 51.80 | 75.20 | 52.48 | 75.90 |
| 3.25 | 100 | 100 | 100 | 31.85 | 3.30 | 32.80 | 65.93 | 51.65 | 66.40 |
| 3.50 | 100 | 100 | 100 | 19.85 | 2.15 | 20.35 | 59.93 | 51.08 | 60.18 |

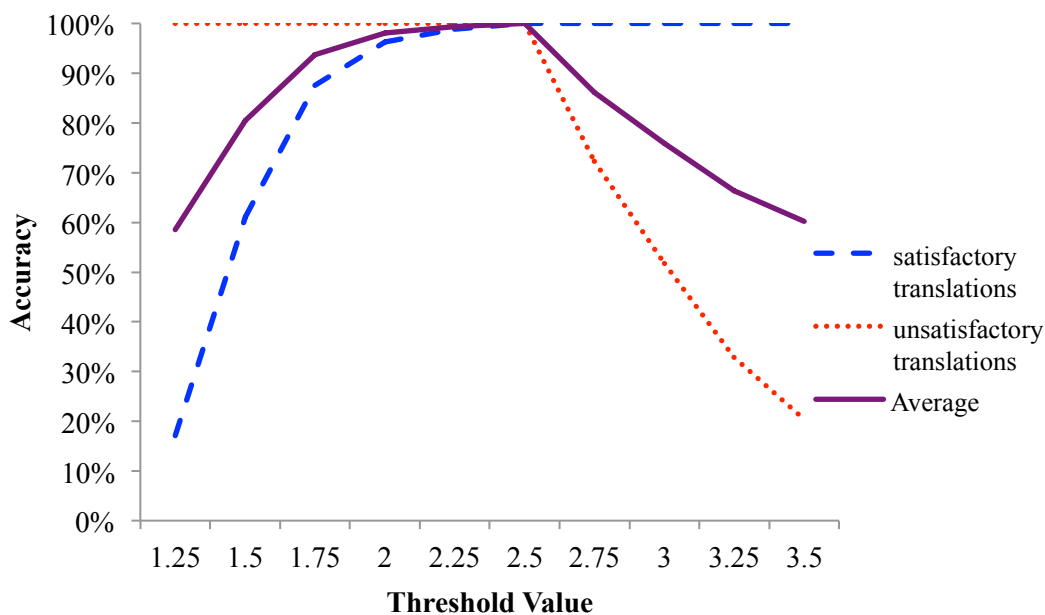

Figure 1. Tendencies of the classification of satisfactory translations and unsatisfactory translations for test Corpus B with different threshold values.

A further experiment was conducted to investigate whether different threshold values are more effective when using the combined *SLR&CR* technique. Table 9 displays the accuracy results matrix of the experiments on the overall accuracy averages on the same 10000 satisfactory translations and 2000 unsatisfactory translations in test Corpus B used in the previous experiment. In the table, the *SLR* threshold value is shown across the top row, and the *CR* threshold value is shown down the left column, both ranging from 1.25 up to 3.50. The table shows that 100% accuracy is achieved using threshold values 2.50 and higher for *SLR* combined with 2.25 and higher for *CR*.

Table 9. The accuracy results matrix for test Corpus B using threshold values of *SLR* and *CR* from 1.25 to 3.50.

| *SLR* /*CR* | 1.25 (%) | 1.50 (%) | 1.75 (%) | 2.00 (%) | 2.25 (%) | 2.50 (%) | 2.75 (%) | 3.00 (%) | 3.25 (%) | 3.50 (%) |
|---|---|---|---|---|---|---|---|---|---|---|
| **1.25** | 17.11 | 58.01 | 82.33 | 88.42 | 89.62 | 89.91 | 89.91 | 89.91 | 89.91 | 89.91 |
| **1.50** | 19.58 | 61.10 | 86.64 | 94.81 | 97.05 | 97.86 | 97.86 | 97.86 | 97.86 | 97.86 |
| **1.75** | 20.13 | 61.95 | 87.58 | 95.82 | 98.17 | 99.24 | 99.24 | 99.24 | 99.24 | 99.24 |
| **2.00** | 20.27 | 62.29 | 88.00 | 96.30 | 98.66 | 99.76 | 99.76 | 99.76 | 99.76 | 99.76 |
| **2.25** | 20.29 | 62.35 | 88.15 | 96.50 | 98.90 | 100 | 100 | 100 | 100 | 100 |
| **2.50** | 20.29 | 62.35 | 88.15 | 96.50 | 98.90 | 100 | 100 | 100 | 100 | 100 |
| **2.75** | 20.29 | 62.35 | 88.15 | 96.50 | 98.90 | 100 | 100 | 100 | 100 | 100 |
| **3.00** | 20.29 | 62.35 | 88.15 | 96.50 | 98.90 | 100 | 100 | 100 | 100 | 100 |
| **3.25** | 20.29 | 62.35 | 88.15 | 96.50 | 98.90 | 100 | 100 | 100 | 100 | 100 |
| **3.50** | 20.29 | 62.35 | 88.15 | 96.50 | 98.90 | 100 | 100 | 100 | 100 | 100 |

Another experiment was devised to determine how much of the larger test Corpus A would be classified as satisfactory or unsatisfactory using various *CR* threshold values (from 1.25 to 3.50) when the *SLR* threshold value was set at 2.5. The results of this experiment are shown in Table 10. The table shows the number classified in each category (in the columns labelled "Amount") and the corresponding percentages. For example, a threshold value of 2.50 for both *SLR* and *CR* results in 8.18% of test Corpus A being labelled unsatisfactory (and therefore candidates for being removed from the corpus).

Table 10: Percentages of satisfactory and unsatisfactory translations for test Corpus A when the *SLR* threshold is set at 2.5.

| Threshold *CR* | Satisfactory Translations | | Unsatisfactory Translations | |
|---|---|---|---|---|
| | Amount | Percentage (%) | Amount | Percentage (%) |
| **1.25** | 1313387 | 72.14 | 507234 | 27.86 |
| **1.50** | 1559275 | 85.65 | 261346 | 14.35 |
| **1.75** | 1626973 | 89.36 | 193648 | 10.64 |
| **2.00** | 1650374 | 90.65 | 170247 | 9.35 |
| **2.25** | 1665709 | 91.49 | 154912 | 8.51 |
| **2.50** | 1671768 | 91.82 | 148853 | 8.18 |
| **2.75** | 1674677 | 91.98 | 145944 | 8.02 |
| **3.00** | 1675700 | 92.04 | 144921 | 7.96 |
| **3.25** | 1676166 | 92.07 | 144455 | 7.93 |
| **3.50** | 1676311 | 92.07 | 144310 | 7.93 |

Figures 2, 3, 4 and 5 show correlations for the sentence length and code length metrics for test Corpus A. Figures 2 and 3 illustrate the sentence lengths and code lengths of Arabic and

English sentences classified as unsatisfactory for the test Corpus A and show an obvious split in the plot due to 1:2 and 2:1 type mismatches.

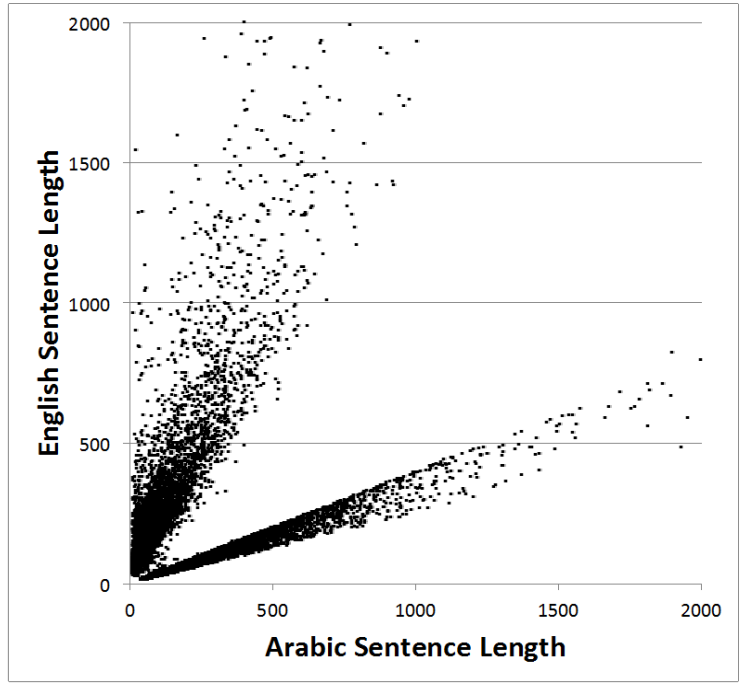

Figure 2. Sentence length correlation for test Corpus A for sentence pairs classified as unsatisfactory.

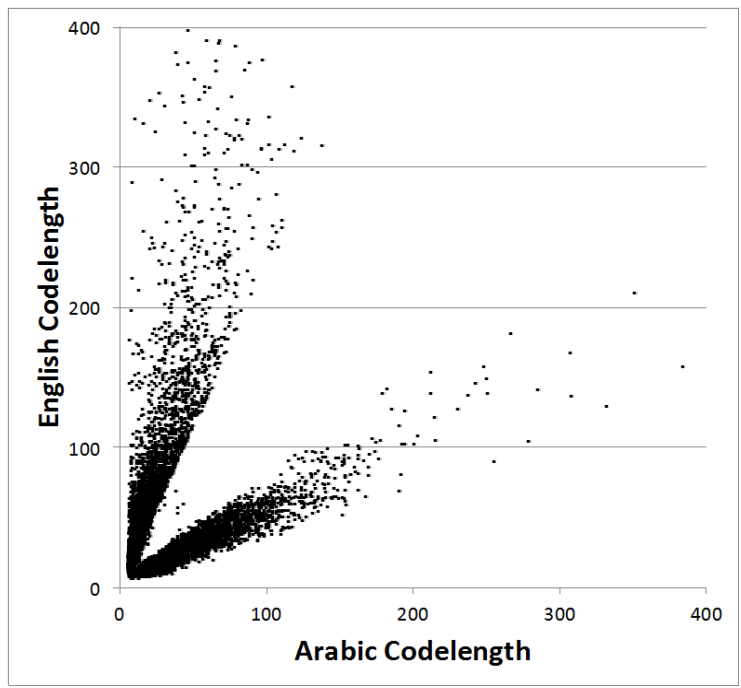

Figure 3. Code length correlation for test Corpus A for sentence pairs classified as unsatisfactory.

In contrast, Figures 4 and 5 illustrate sentence lengths and code lengths of Arabic and English for the translations classified as satisfactory for test Corpus A and show a strong correlation between both sentence lengths and compression code lengths.

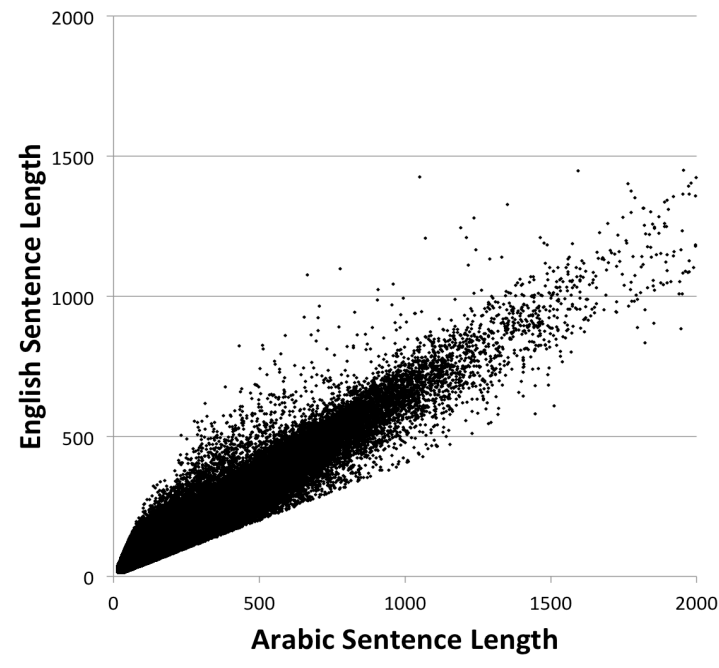

Figure 4. Sentence length correlation for test Corpus A for translations classified as satisfactory.

For defining what is a satisfactory translation in this case, it was decided if the values of *SLR* were less than 2.5 and *CR* less than 2.25 for a pair of translation sentences, then it is classified as a satisfactory translation, otherwise it is classified as an unsatisfactory translation as per Figures 2 and 3.

The unsatisfactory translations might be caused by errors in alignment between Arabic and English sentences which may include non-literal translations and therefore result in significant differences between the sentence pair. English sentences containing websites or abbreviations such as USA (United States of America), UNCTAD (United Nations Conference on Trade and Development Abbreviation) might also lead to mistranslations [10].

The flowchart in Figure 6 shows how the new hybrid metric was applied in this manner. If the *CR* threshold of 2.25 was exceeded, or the *SLR* threshold of 2.5 was exceeded, then the sentence pair would be rejected (i.e. classified as unsatisfactory), otherwise the translation was accepted (i.e. classified as satisfactory).

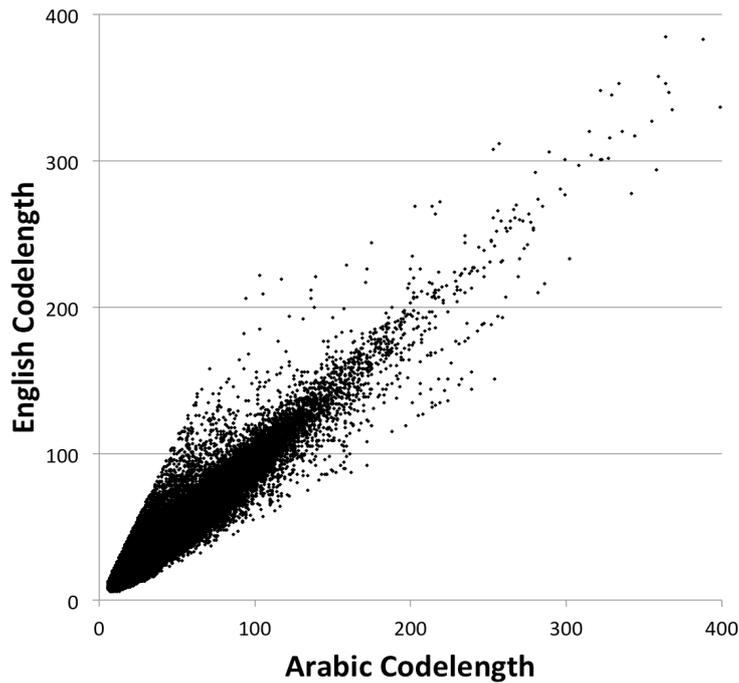

Figure 5. Code length correlation for test Corpus A for translations classified as satisfactory.

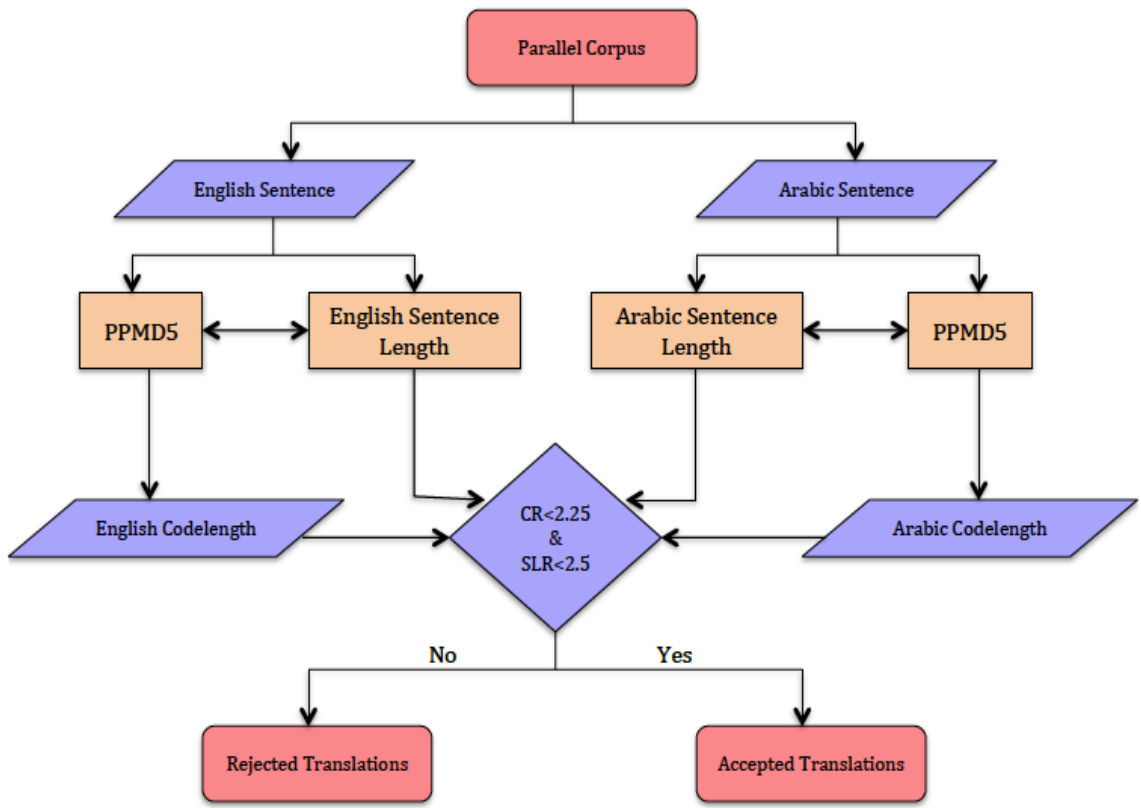

Figure 6. Flow chart of how the new hybrid sentence matching metric based on both compression code length and sentence length was applied to test Corpus A.

## 5. CONCLUSION

Verification is an essential step in order to ensure a high quality corpus. In this paper, we have described a new method to check how well the sentences match in a parallel corpus. The method is based on the combination of two distance metrics, sentence length ratio (*SLR*) and compression code length ratio (*CR*). A threshold mechanism can be used to filter out unsatisfactory translations when either the *SLR* or *CR* values have been exceeded. Experiments with a small sample of sentence pairs from a test Arabic-English corpus containing ground truth judgments, which were manually judged to be satisfactory or unsatisfactory translations, show that a combination of both *SLR* and *CR* distance metrics performs better at classification than using a single distance metric by itself.

There is also other important verification tasks that are often overlooked not described here that need to be done. For example, a single check on document sizes is crucial (e.g. ensuring no zero byte documents, and removing unusually large documents if appropriate). Checking for self-plagiarism (ensuring that documents do not contain strings repeated in other documents) is also essential (especially for corpora containing news stories since it is a common practice for these types of documents to contain material copied from other news stories). We have found that the compression code length metric described here is also effective at classifying the quality of translation not just at the sentence level, but also at the document, paragraph and clause levels, and these should also be checked when verifying a parallel corpus.

## REFERENCES

[1] Alhawiti, K., (2014) "Adaptive Models of Arabic Text", *PhD Dissertation, Bangor University*.

[2] Behr F. H., Fossum V., Mitzenmacher M., Xiao D., (2003) "Estimating and Comparing Entropy across Written Natural Languages Using PPM Compression", *Proceedings of Data Compression Conference*, p416.

[3] Brown, P., Della Pieta, S., Della Pieta, V., Mercer, R., (1993) "The Mathematics of Machine Translation: Parameter Estimation", *Computational Linguistics*, Vol. 19, pp263-312.

[4] Cleary, J. G. & Witten, I. H., (1984) "Data Compression Using Adaptive Coding and Partial String Matching", *IEEE Transactions on Communications*, Vol. 32, No. 4, pp396-402.

[5] Fantechi, A., Gnesi, S., Carenini, M., Vanocchi, M., Moreschini, P., (1994) "Assisting Requirement Formalization by Means of Natural Language Translation", *Formal Methods in System Design*, Vol. 4, No. 3, pp243-263.

[6] Gale, W.A. & Church, K.W., (1993) "A Program for Aligning Sentences in Bilingual Corpora", *ACL'93 29th Annual Meeting*, pp177-184.

[7] Haruno, M. & Yamazaki, T., (1996) "High-performance Bilingual Text Alignment Using Statistical and Dictionary Information", *Proceedings of the 34th Annual Meeting of Association for Computational Linguistics*, pp131-138.

[8] Hutchins, W. J., (1994) "The Encyclopaedia of Languages and Linguistics", *ed. R. E. Asher, Oxford: Pergamon Press*, Vol. 5, pp2322-2332.

[9] Kay, M. & Röscheisen, M., (1993) "Text-translation Alignment", *Computational Linguistics*, Vol. 19, pp121-142.

[10] Khadivi, S. & Ney, H., (2005) "Automatic Filtering of Bilingual Corpora for Statistical Machine Translation", *Natural Language Processing and Information Systems*, Vol. 3513, pp263-274.

[11] Kutuzov, A., (2013) "Improving English-Russian Sentence Alignment through POS Tagging and Damerau-Levenshtein Distance", *Association for Computational Linguistics*, pp63-68.

[12] Linguistic Data Consortium, *http://catalog.ldc.upenn.edu*

[13]   Liu, W., Chang, Z., Teahan, W., (2014) "Experiments with Compression-based Methods for English-Chinese Sentence Alignment", *2nd International Conference on Statistical Language and Speech Processing*, pp70-81.

[14]   Melamed, I. D., (2000) "Models of Translational Equivalence among Words", *Computational Linguistics*, Vol. 26, No. 2, pp221-249.

[15]   Mubarak, H., Darwish, K., Adly, N., (2014) "Using Twitter to Collect a Multi-Dialectal Corpus of Arabic", *EMNLP 2014 Workshop on Arabic Natural Language Processing*.

[16]   Mújdricza-Maydt, É., Körkel-Qu, H., Riezler, S., Padó, S., (2013) "High-Precision Sentence Alignment by Bootstrapping from Wood Standard Annotations", *The Prague Bulletin of Mathematical Linguistics*, Vol. 99, pp5-16.

[17]   Papageorgiou, H., Cranias, L., Piperidis, S., (1994) "Automatic Alignment in Corpora", *Proceedings of 32nd Annual Meeting of Association of Computational Linguistic*, pp334-336.

[18]   Shannon, C. E., (1948) "A Mathematical Theory of Communication", *Bell System Technical Journal*, Vol. 27, pp379-423 & pp623-656.

[19]   Simard, M., Foster, G. F., Isabelle, P., (1992) "Using Cognates to Align Sentences in Bilingual Corpora", *Proceedings of the Fourth International Conference on Theoretical and Methodological Issues in Machine Translation (TMI)*, pp67–81.

[20]   Teahan, W., (1998) "Modelling English Text", *PhD Dissertation, University of Waikato, New Zealand*.

[21]   Wu, D., (1994) "Aligning a Parallel English-Chinese Corpus Statistically with Lexical Criteria", *ACL'94 32nd Annual Meeting*, pp80-87.

[22]   Yu, Q., Max, A., Yvon, F., (2012) "Revisiting Sentence Alignment Algorithms for Alignment Visualization and Evaluation", *LREC Workshop*, pp10-16.